\documentclass[runningheads]{llncs}

\usepackage[T1]{fontenc}
\usepackage{soul}
\usepackage{comment}
\usepackage{graphicx}
\usepackage{amsmath}
\usepackage{bbm}
\usepackage{amssymb,enumitem}
\usepackage{amsmath, color, amssymb}
\usepackage{algorithm}
\usepackage{algpseudocode}
\usepackage{float}
\usepackage{xcolor}
\usepackage{amsfonts}
\usepackage{graphicx}
\usepackage{multirow}
\usepackage{comment}
\usepackage{caption}
\usepackage{hyperref}
\usepackage{orcidlink}

\begin{document}

\title{\textit{MIXAD}: Memory-Induced Explainable Time Series Anomaly Detection}

\author{Minha Kim\inst{1}\orcidlink{0000-0002-3224-0610} \and
Kishor Kumar Bhaumik\inst{1,2}\orcidlink{0009-0004-8088-2492} \and
Amin Ahsan Ali \inst{2}\orcidlink{2222--3333-4444-5555} \and
Simon S.  Woo\inst{1}\thanks{Corresponding Author}\orcidlink{0000-0002-8983-1542}}
\authorrunning{Kim et al.}

\institute{Sungkyunkwan University, Republic of Korea \\
\email{\{sunshine01, kishor25, swoo\}@g.skku.edu} \and
Center for Computational \& Data Sciences, Independent University, Bangladesh \\
\email{aminali@iub.edu.bd}}

\maketitle             

\begin{abstract}
For modern industrial applications, accurately detecting and diagnosing anomalies in multivariate time series data is essential. Despite such need, most state-of-the-art methods often prioritize detection performance over model interpretability. Addressing this gap, we introduce \textbf{\textit{MIXAD}} (\underline{\textbf{M}}emory-\underline{\textbf{I}}nduced E\underline{\textbf{x}}plainable Time Series \underline{\textbf{A}}nomaly  \underline{\textbf{D}}etection), a model designed for interpretable anomaly detection. \textit{MIXAD} leverages a memory network alongside spatiotemporal processing units to understand the intricate dynamics and topological structures inherent in sensor relationships. We also introduce a novel anomaly scoring method that detects significant shifts in memory activation patterns during anomalies. Our approach not only ensures decent detection performance but also outperforms state-of-the-art baselines by \textbf{34.30\%} and \textbf{34.51\%} in interpretability metrics. The code for our model is available at \url{https://github.com/mhkim9714/MIXAD}.
\end{abstract}

\section{Introduction}
The proliferation of sensors and Internet of Things (IoT) devices in healthcare, smart manufacturing, and cybersecurity has significantly increased the generation of time series data, emphasizing the need for advanced Multivariate Time Series (MTS) analysis~\cite{chalapathy2019deep,cook2019anomaly,yg}. Anomaly detection, crucial for system diagnosis and maintenance, faces challenges due to the rarity of anomalies and the dynamic nature of data. These issues complicate manual labeling and necessitate a deep understanding of both intra-metric (within a single series) and inter-metric (across different series) dependencies. Furthermore, enhancing explainability in MTS anomaly detection is essential for improving operational clarity and decision-making, particularly in pinpointing and explaining the root causes of anomalies. In environments like smart manufacturing with numerous sensors, clarity about which sensors or interactions between sensors contribute to an anomaly is vital. Despite existing methods' capabilities in detecting anomalies, many lack the ability to provide clear explanations for their findings~\cite{kieu2019outlier}. Enhancing autoencoder models with an external memory module has emerged as a promising solution. This approach not only boosts detection performance but also deepens insights into the dynamics of anomalies, thereby improving model interpretability and building trust in anomaly detection systems~\cite{hutchison2022explainable}.

Addressing gaps in MTS anomaly detection, we introduce \textit{MIXAD}. For the first time in MTS anomaly detection, \textit{MIXAD} incorporates a memory module to store the complex dynamics of data extracted through a spatiotemporal feature extractor. This adoption sets \textit{MIXAD} apart from most baselines, as our model simultaneously models intra-metric and inter-metric dependencies, offering enhanced insight into the nature of the data. Additionally, we generate a novel anomaly score through memory activation pattern analysis, which successfully combines the accuracy of anomaly detection with improved elucidation of root causes. Unlike most existing methods that rely heavily on forecasting or reconstruction errors to construct the anomaly score—typically identifying the feature with the largest error as the cause of the anomaly—our approach employs a more sophisticated scoring method. This method explains which features contribute to the anomaly and how by analyzing the differences in memory activation patterns between normal and abnormal periods. Furthermore, by adopting the Pearson correlation calculation in a post hoc manner, we can further determine which features share similar patterns of memory activation shift. \textit{MIXAD} offers critical, actionable insights for practical applications by blending the precision of unsupervised learning with the transparency of explainable AI. The contributions of this work are outlined as follows:
\begin{enumerate}
    \item We present \textit{MIXAD}, a pioneering approach that significantly improves the interpretability of anomaly detection outcomes in the MTS domain. This innovation addresses a significant oversight in existing research, making the model applicable across diverse fields.
    \item For the first time in MTS anomaly detection, \textit{MIXAD} integrates a memory network designed to capture and retain the normal spatiotemporal patterns of data. We also introduce a novel anomaly scoring mechanism that leverages memory attention pattern matching, maintaining robust detection performance while significantly augmenting the model's interpretive capabilities.
    \item We employ benchmark datasets with interpretation labels, enabling thorough evaluation of our model's detection capabilities and interpretive efficacy. \textit{MIXAD} not only achieves robust detection results but also significantly outperforms existing state-of-the-art (SOTA) baselines in interpretability metrics by \textbf{34.30\%} and \textbf{34.51\%}.
\end{enumerate}

\section{Related Works}
\textbf{Time Series Anomaly Detection. }Anomaly detection in large-scale databases has become more challenging with the emergence of diverse data modalities~\cite{he2021automl,kieu2019outlier}. The widespread use of sensors and IoT devices has significantly increased data volumes, underscoring the need for precise anomaly detection in time-series databases. These databases often exhibit stochastic and temporal patterns from various engineering sources, necessitating effective differentiation of outliers~\cite{OmniAnomaly}. Due to the scarcity of labeled data and the diversity of anomalies~\cite{chalapathy2019deep}, the current research trend is biased towards unsupervised learning models. Time-series anomaly detection research is broadly classified into two main categories: univariate models that analyze individual time series and multivariate methods that examine multiple series concurrently. Despite progress, many SOTA methods still lack a quantitative evaluation of model explainability, which is essential for real-world applications. Our work focuses on enhancing the interpretability of models in practical environments. For detailed discussions on recent developments, please refer to the supplementary materials.

\noindent \textbf{Memory Network for Anomaly Detection. }Recent advancements in anomaly detection have highlighted the effectiveness of memory-augmented attention (MAA) models, particularly for their capacity to store normal data patterns encountered during training. For example, Gong et al.\cite{gong2019memorizing} integrated a memory module into an autoencoder to mitigate the model's tendency to reconstruct anomalies accurately, thus enhancing detection precision using reconstruction errors. Similarly, Park et al.\cite{park2020learning} developed an unsupervised video anomaly detection method that utilizes a memory module to record normal data patterns, enhancing the discriminative power of the memory items and the abnormal features. This approach also uses feature compactness and separateness losses to ensure the diversity and discriminative capabilities of memory items, thereby improving anomaly detection efficiency and effectiveness. Inspired by these approaches, we propose the novel integration of a memory module into the MTS anomaly detector to enhance performance, providing deeper insights into how anomalous patterns diverge from normal patterns.

\begin{figure*}[!t]
 \centering
 \includegraphics[width=1\textwidth]{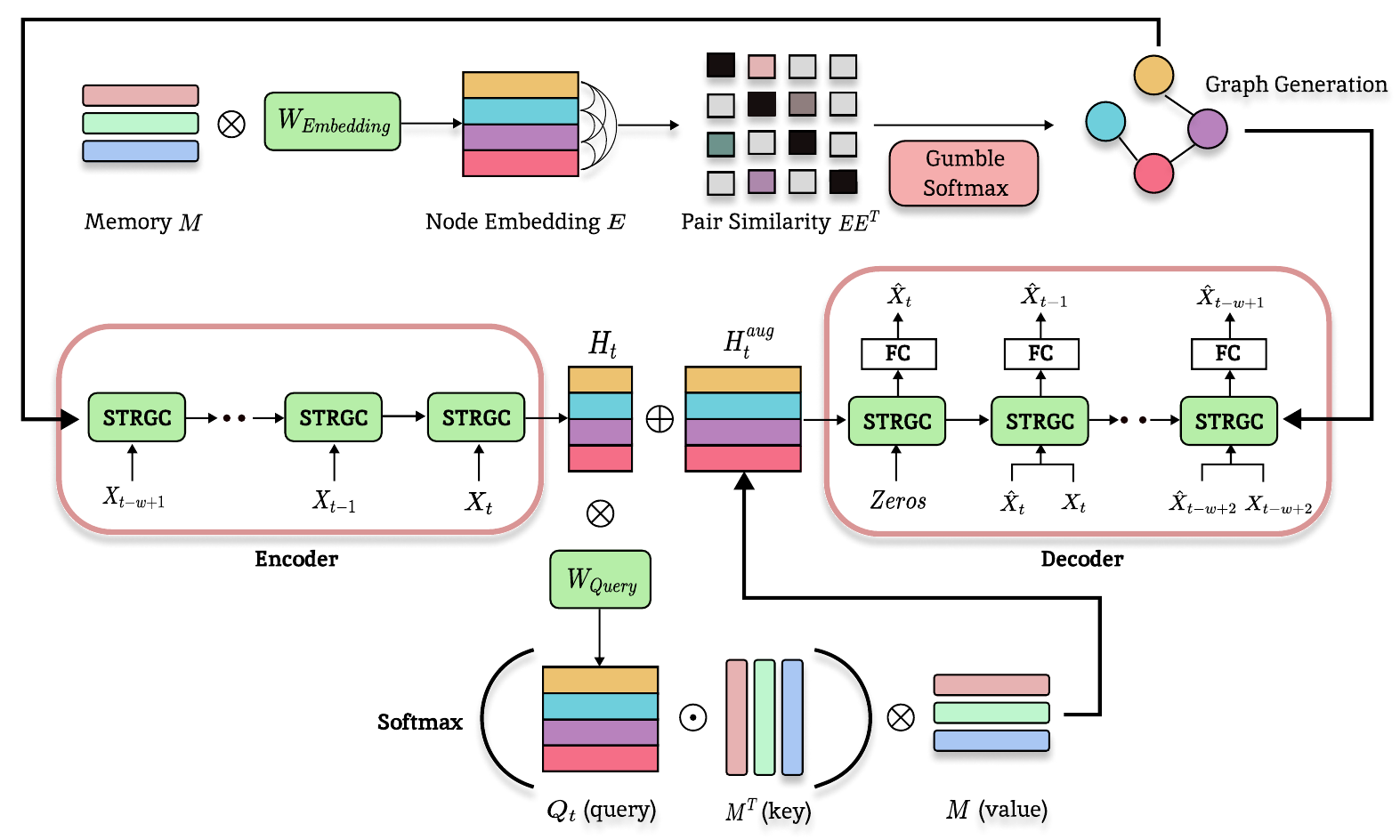}
 \caption{Overview of the \textit{MIXAD} framework: Initially, a sparse graph is constructed by calculating pairwise similarities between memory-based node embeddings. Subsequently, input data is processed through a STRGC-based encoder and decoder for self-reconstruction. Throughout this process, an external memory enhances the encoded feature vector by utilizing an attention mechanism between the original feature vector and the memory.}
 \label{fig:framework}
\end{figure*}

\section{Method}
Figure~\ref{fig:framework} provides an overview of the \textit{MIXAD} framework. The model features an encoder-decoder architecture, with the Spatiotemporal Recurrent Convolution Unit (STRGC) serving as the core component for both the encoder and the decoder. An external memory assists in augmenting the encoded hidden representations to better initialize the decoder's hidden states. Additionally, the learned memory is processed through an embedding layer to generate the graph structure utilized within the STRGC. The subsequent subsections will detail each component of \textit{MIXAD}, beginning with an explanation of the basic problem formulation.

\subsection{Problem Formulation}
In this study, we introduce a framework for unsupervised anomaly detection, analyzing time series data with $N$ features over a period $T$, denoted as $X \in \mathbb{R}^{N \times T}$. For simplicity, we denote feature indices with superscripts and timestamps with subscripts, allowing for the representation of data at any time $t$ as $X_t \in \mathbb{R}^{N \times 1}$. We normalize both training and testing data using feature-wise min-max normalization and employ a sliding window technique with window size $w$ to capture temporal dependencies, denoted as $W_t = X_{t-w+1:t}$. As the traditional reconstruction-based methods, our proposed model, $f$, is trained to reconstruct the input window: $\hat{W_t} = f(W_t)$. However, \textit{MIXAD} distinguishes itself from existing methods by adopting a novel anomaly scoring mechanism that utilizes memory activation analysis. The model computes an anomaly score $s_t$ for each timestamp $t$ and compares it to a predefined threshold to determine anomalies. Since \textit{MIXAD} aims to not only detect anomalies but also pinpoint the specific features contributing to anomalies, our main objectives can be summarized as follows:
\begin{itemize}
    \item \textbf{Anomaly Detection: } Determine if $X_t$ at a given timestamp is anomalous.
    \item \textbf{Anomaly Interpretation:} Identify which features contribute to the anomaly at the identified timestamps.
\end{itemize}

\subsection{Spatiotemporal Recurrent Convolution Unit}
Drawing inspiration from the successful implementation of temporal modeling with LSTM-based encoder-decoder structures~\cite{EncDec-AD}, \textit{MIXAD} adopts a similar architecture. The encoder first processes the input time series window, extracting a fixed-length feature vector. Subsequently, the decoder utilizes this representation to reconstruct the time series in reverse order. 

Recognizing that MTS exhibit both temporal and spatial dependencies among features, recent advancements in GNNs have led to a notable development of STRGC, which incorporates graph convolution operations into recurrent cells~\cite{MegaCRN,RGSL}. In our work, by substituting LSTM with STRGC, \textit{MIXAD} concurrently processes all series, capturing the data's spatial and temporal dependencies more effectively. The graph convolution operation is defined as follows:
\begin{gather}
\label{eqn:eqn2}
    W_{\star A}X_t = \sum_{k=0}^{K} \Tilde{A}^k X_t W_k
\end{gather}
where $_{\star A}$ represents a graph convolution operation with input $X_t \in \mathbb{R}^{N \times 1}$ and kernel parameters $W \in \mathbb{R}^{K \times 1 \times h}$, approximated using Chebyshev polynomials to the order of $K$~\cite{chebyshev}. This operation requires an adjacency matrix $A \in \mathbb{R}^{N \times N}$, normalized to $\Tilde{A}$, to outline the data's topological structure. Building on this, the STRGC-enhanced GRU cell updates are as follows:
\begin{equation}
\label{eqn:eqn3}
\begin{aligned}
    r_t &= sigmoid(W_{r \star A} [X_t \mathbin\Vert H_{t-1}] + b_r) \\
    u_t &= sigmoid(W_{u \star A} [X_t \mathbin\Vert H_{t-1}] + b_u) \\
    C_t &= tanh(W_{C \star A} [X_t \mathbin\Vert (r_t \odot H_{t-1})] + b_C) \\
    H_t &= u_t \odot H_{t-1} + (1-u_t) \odot C_t
\end{aligned}
\end{equation}
where $r$, $u$, and $C$ denote the reset gate, update gate, and candidate state within the GRU cell, respectively, with $W_{\{r,u,C\}} \in \mathbb{R}^{K \times (1+h) \times h}$ and $b_{\{r,u,C\}} \in \mathbb{R}^h$ representing the gate parameters. The hidden representation produced by STRGC is denoted by $H_t \in \mathbb{R}^{N \times h}$. This adaptation enables our model to leverage graph convolution within a GRU cell, providing a sophisticated method to dissect spatiotemporal dependencies in multivariate time series data.

\subsection{Memory-Augmented Graph Structure Learning}
In the realm of spatiotemporal modeling, tasks like traffic forecasting leverage spatial adjacency information, often readily available from urban maps~\cite{RGSL}. However, applying similar spatial correlation concepts to MTS anomaly detection presents challenges. In MTS anomaly detection, identifying dependencies among features is not immediately obvious and typically requires domain expertise to formalize these correlations. To overcome this obstacle, our research incorporates a Graph Structure Learning (GSL) method. GSL excels at systematically discovering and applying spatial correlations among features without prior explicit knowledge, thereby significantly enhancing the anomaly detection capabilities of models dealing with complex MTS data.

To streamline our examination of GSL within the context of MTS anomaly detection, we highlight its application in the Graph Deviation Network (GDN) framework~\cite{GDN}. GDN employs learnable embeddings for each feature, establishing feature connections based on embedding similarity. While effective, this method creates a static graph that may not fully accommodate the dynamic nature of non-stationary time series data. To address these limitations and accommodate the temporal evolution of data, we employ a dynamic graph learning strategy introduced in Structure Learning Convolution (SLC)~\cite{SLC}. SLC utilizes input-conditioned node embeddings, creating a graph that dynamically adapts over time, thus providing a more flexible and responsive modeling of feature relationships. This method enhances our model's ability to adjust to changing data patterns, although challenges remain in preventing over-sensitivity to immediate data variations and ensuring consistency in the graph structure.

To address the aforementioned limitations of traditional GSL methods, we draw from foundational research~\cite{MegaCRN} to enhance spatiotemporal graph learning with a memory module. In \textit{MIXAD}, each memory item is updated at each iteration, storing node-level spatiotemporal prototypes that generate node embeddings adaptable to non-stationary time series without being overly sensitive to noisy data. The memory module, denoted as $M \in \mathbb{R}^{m \times d}$, where $m$ represents the number of memory items and $d$ the dimension of each item, plays a pivotal role in pattern recognition and modeling feature relationships. The interaction with the memory module is captured through:
\begin{equation}
\label{eqn:eqn6}
\begin{aligned}
     &Q_t = H_t * W_Q + b_Q \\
     &Att_t = softmax(Q_t * M^T) \\
     &H_t^{aug} = Att_t * M
\end{aligned}
\end{equation}
Here, $H_t^{aug} \in \mathbb{R}^{N \times d}$ represents a memory-augmented hidden state derived via an attention mechanism~\cite{attention}. The process begins by transforming the hidden representations $H_t$ into a query space, resulting in $Q_t \in \mathbb{R}^{N \times d}$, using the parameters $W_Q \in \mathbb{R}^{h \times d}$ and $b_Q \in \mathbb{R}^d$. The memory module $M$ is then utilized as both key and value in the attention operation, allowing each feature's query vector to compute similarity scores across  $m$ memory items. These attention scores $Att_t \in \mathbb{R}^{N \times m}$ generate an augmented hidden state $H_t^{aug}$ through a weighted aggregation of memory items, enhancing the initial hidden representations. This augmented state is subsequently concatenated with $H_t$, initializing the decoder's hidden state. This approach not only enriches the model's capacity to discern complex feature relationships but also adapts to their dynamism over time.

In our spatiotemporal graph learning framework, the memory module $M$ is essential for creating advanced node embeddings that are crucial for structuring the input for the STRGC encoder and decoder. However, we encountered a challenge with the graph adjacency matrix becoming overly dense, cluttered with unnecessary edges that could obscure the spatial relationships between features. To address this issue, we introduced a regularized graph generation module aimed at promoting sparsity in the adjacency matrix. This ensures that the graph structure more accurately captures the essential feature connections~\cite{RGSL}. The regularized graph generation process can be expressed as follows:
\begin{equation}
\label{eqn:eqn7_2}
\begin{gathered}
    \theta = (W_E * M){(W_E * M)}^T \\
    A = \sigma((log(\theta^{ij}/(1-\theta^{ij}))+({g1}^{ij}-{g2}^{ij}))/\tau) \\
    s.t.\quad {g1}^{ij}, {g2}^{ij} \sim Gumbel(0,1)
\end{gathered}
\end{equation}
where $W_E \in \mathbb{R}^{N \times m}$ projects the memory onto an embedding space, resulting in a probability matrix $\theta \in \mathbb{R}^{N \times N}$. Each element $\theta^{ij}$ reflects the potential for an edge between features $i$ and $j$. Utilizing the Gumbel softmax technique, we convert $\theta$ into a discretely sparse adjacency matrix $A \in \mathbb{R}^{N \times N}$, maintaining differentiability for gradient-based optimization. In Equation~\eqref{eqn:eqn7_2}, $\sigma$ represents the activation function and $\tau$ the softmax temperature.

\subsection{Memory-Induced Explainable Anomaly Detection (MIXAD)}
Our proposed \textit{MIXAD} architecture illustrated in Figure~\ref{fig:framework}, employs the STRGC framework, enhanced with a memory module, to classify nodes with similar spatiotemporal dynamics into specific memory slots. This setup aims to efficiently identify and store distinct patterns within each memory item, improving the model's discriminative power. To refine this capability, we incorporate three specialized loss functions, as follows:

\begin{equation}
\begin{gathered}
\label{eqn:eqn9}
    L_1 = \sum_{t,i}^{T,N} max\{ { \mathbin\Vert Q_t^i - M[pos_t^i] \mathbin\Vert }^2 - { \mathbin\Vert Q_t^i - M[neg_t^i] \mathbin\Vert }^2 + \lambda, 0 \} \\
    L_2 = \sum_{t,i}^{T,N} { \mathbin\Vert Q_t^i - M[pos_t^i] \mathbin\Vert }^2 \\
    L_3 = -log(m) - \frac{1}{m} \sum_{j=1}^m log(\frac{exp(\sum_{t,i}^{T,N} Att_t^{ij})}{\sum_{k=1}^m exp(\sum_{t,i}^{T,N} Att_t^{ik})}) \\
    Loss = L_{MAE} + \lambda_1L_1 + \lambda_2L_2 + \lambda_3L_3
\end{gathered}
\end{equation}

The core of our method leverages $Q_t^i \in \mathbb{R}^d$ as an anchor to identify the nearest ($M[pos_t^i] \in \mathbb{R}^d$) and the second-nearest memory items ($M[neg_t^i] \in \mathbb{R}^d$) based on attention scores. We use a triplet margin loss $L_1$ to ensure a significant margin between the closest and second-closest slots for improved separation, while a compact loss $L_2$ maintains close proximity between the closest slot and the anchor. To mitigate training instabilities and uneven memory utilization, we also implement a Kullback-Leibler (KL) divergence loss $L_3$, which promotes a uniform attention distribution across $m$ memory items. The overall training objective, as presented in the bottom of Equation~\eqref{eqn:eqn9}, combines these losses with a Mean Absolute Error (MAE)-based reconstruction loss, regulated by balancing parameters $\lambda_1$, $\lambda_2$, and $\lambda_3$. This integrated objective stabilizes the training process and enhances the interpretability and efficiency of memory items in our anomaly detection model. We selected MAE as the most suitable reconstruction loss over Mean Squared Error (MSE) and Root Mean Squared Error (RMSE) after considering their characteristics. In unsupervised anomaly detection, normal instances are typically used for training, but high-purity normal datasets are rare in real-world data, making it crucial to minimize the influence of noise. MSE and RMSE are significantly affected by noise due to squaring the differences between predicted and ground truth values, leading to training instability. Therefore, we adopted the more robust MAE loss.


\subsection{Anomaly Scoring}
Traditional reconstruction-based anomaly detectors utilize the reconstruction error to generate an anomaly score, as shown by the equation $s_t = |W_t - \hat{W_t}|$. However, advanced models often reconstruct anomalies too accurately, thus diminishing the effectiveness of this approach. Furthermore, the presence of contextual anomalies, where individual series distributions appear normal, but their interrelations are anomalous, necessitates more sophisticated scoring methods.

To address these challenges, we introduce a novel anomaly scoring method that leverages memory query attention scores. This method assumes that inputs deviating from learned normal patterns will significantly alter the distribution of attention scores for the memory components. The anomaly score for timestamp $t$, $s_t \in \mathbb{R}^N$, is calculated by evaluating the shift in attention score distributions between consecutive timestamps, using the Jensen-Shannon divergence (JSD):
\begin{gather}
\label{eqn:eqn10}
    s_t = [JSD(Att_{t-1}^i \mathbin\Vert Att_t^i)]_{i=1}^N
\end{gather}

Subtle differences in memory activation patterns, even when the reconstruction error is not notably large or in the presence of contextual anomalies, enable the model to identify anomalies more accurately. Furthermore, understanding how memory activation changes enhances our ability to interpret the detected anomalies deeply. For example, if a node that typically references memory item 1 shifts to referencing memory item 2 at a certain timestamp, obtaining this information provides a deeper insight into the nature of the anomaly. Meanwhile, for time series that display distinct cyclic temporal patterns, normal seasonal fluctuations might inadvertently affect the anomaly score $s \in \mathbb{R}^{T \times N}$, complicating anomaly detection. To mitigate this, we apply Seasonal-Trend decomposition using LOESS (STL)~\cite{STL} on the anomaly scores and remove identifiable seasonal components based on the period $P$, determined from the Real Fast Fourier Transform (RFFT) analysis~\cite{RFFT} of $s^i$ for each feature $i$. The process is defined as:
\begin{equation}
\label{eqn:eqn11}
\begin{gathered}
    \forall i \in \{1,...,N\}, {s'}^i = s^i - STL\{s^i;P\}_{seasonal} \\
    P = (\frac{2 \pi}{\Delta t \cdot argmax(|\mathcal{F}\{s^i\}|)})
\end{gathered}
\end{equation}
Using the dominant frequency of signal $s^i$ identified by the maximum modulus of the FFT ($\mathcal{F}$) output, $P$ is calculated from the RFFT frequency spectrum, with $\Delta t$ as the sampling interval. The seasonal component is extracted using STL, and subsequent deseasonalization yields a refined anomaly score $s' \in \mathbb{R}^{T \times N}$. To determine the overall anomaly level at each timestamp $t$, we aggregate ${s'}_t$ across all features using the max function. A timestamp $t$ is flagged as anomalous if its aggregated score surpasses a set threshold.

\subsection{Anomaly Interpretation}
Anomaly Interpretation is essential in anomaly detection, yet it has been insufficiently addressed in much of the recent research~\cite{DAEMON,Interfusion}. The ability to interpret and trust the outputs of deep learning models is crucial, especially as these models are increasingly utilized to analyze datasets in various real-world applications. Traditionally, identifying the contributing factors of detected anomalies has depended on analyzing the reconstruction or prediction error for each data dimension. However, this approach hinges on the model's accuracy, potentially compromising interpretability when models inaccurately handle input data. Our examination of existing literature reveals a notable gap in quantitatively measured interpretability. To bridge this gap and enhance the utility of anomaly detection, we introduce a new method for anomaly interpretation in our study.

For each detected anomalous segment, we observe that the anomaly scores of ground truth causal features often exhibit similar patterns that set them apart from non-causal features. To pinpoint the set of causal features of an anomaly, we begin by identifying the feature with the highest anomaly score in the segment. We then calculate the Pearson correlation coefficient between the anomaly scores of every feature and the anomaly score of the identified feature. Ranking the features based on the absolute values of their correlation coefficients allows us to order them from most to least likely to have caused the anomaly. This approach provides a detailed and interpretable method for analyzing anomalies, improving the model's utility and reliability in practical applications.

\section{Experiments and Analysis}
\label{section:experiment}

\subsection{Datasets and Baselines}
\textbf{Datasets.}We utilized two publicly available datasets, the Server Machine Dataset (SMD)\cite{OmniAnomaly} and the Multi-Source Distributed System (MSDS) Dataset\cite{MSDS}, specifically chosen for their availability of ground truth information on anomalous features within the datasets. These additional interpretation labels enable us to quantitatively evaluate and compare the effectiveness of our interpretative approach against the baselines. Detailed information about these datasets is provided in the supplementary materials.

\noindent \textbf{Baselines. }We compared a range of selected SOTA anomaly detection algorithms, chosen for their explicit emphasis on enhancing the interpretability of detection results, against our \textit{MIXAD}. The algorithms, MTAD-GAT~\cite{MTAD_GAT}, GDN~\cite{GDN}, TranAD~\cite{TranAD}, DuoGAT~\cite{DuoGAT}, and DAEMON~\cite{DAEMON}, are notable for their advanced methods in elucidating complex data interactions and enhancing detection accuracy. For a comprehensive description of each baseline algorithm, please refer to the supplementary materials.

\subsection{Evaluation Metrics}
To evaluate \textit{MIXAD}'s effectiveness against competing models, we use precision, recall, and the F1-score, incorporating a point-adjusted evaluation method for a more realistic assessment of anomaly detection~\cite{OmniAnomaly,Point_Adjust}. This method acknowledges that real-world anomalies typically span multiple timestamps, treating the identification of any part of an anomaly segment as a correct detection. For anomaly interpretation, \textit{MIXAD}'s ability to detect actual anomalous features among its top predictions is measured using the HitRate@P\% metric. This metric adjusts the evaluation scope based on the proportion of ground truth features, offering a nuanced understanding of the model's diagnostic accuracy. The formula for calculating HitRate@P\% is as follows:
\begin{gather}
\label{eqn:eval0}
    HitRate@P\% = \frac{Hit@ \lfloor P \% \times |GT| \rfloor}{|GT|}
\end{gather}
where $|GT|$ denotes the number of ground truth causal features, and $P\%$ represents the percentage of ground truth dimensions evaluated at each timestamp. Following prior works~\cite{TranAD}, we use 100 and 150 for $P$.

\begin{center}
    \begin{table*}[!t]
\centering
\caption{Performance comparison of \textit{MIXAD} with baseline models on the SMD and MSDS datasets. The highest performance according to the F1-score and HitRate@P\% is highlighted in bold, while the second-best performance is underlined.}
\scalebox{0.9}{
\begin{tabular}{c|cccc||cc}
\hline
\multicolumn{1}{c|}{\textbf{Dataset}} & \multicolumn{1}{c}{\textbf{Method}}       & \multicolumn{1}{c}{\textbf{Precision}} & \multicolumn{1}{c}{\textbf{Recall}} & \multicolumn{1}{c|}{F1}           & \multicolumn{1}{c}{\textbf{HitRate@100\%}}   & \multicolumn{1}{c}{\textbf{HitRate@150\%}}   \\ \hline
                              & MTAD-GAT                          & 0.8889                         & 0.7943                      & 0.8318                            & 0.3716                               & 0.4801                               \\
                              & GDN                               & 0.9114                         & 0.8917                      & 0.8968                            & 0.2994                               & 0.4285                               \\
SMD                           & TranAD                            & 0.9595                         & 0.9325                      & 0.9446                            & 0.3628                               & 0.4747                               \\
                              & DuoGAT                            & 0.9924                         & 0.9945                     & \textbf{0.9965}                   & \underline{0.3825}                         & \underline{0.5155}                         \\
                              & DAEMON                            & 0.9456                         & 0.9746                      & 0.9595                            & 0.3304                               & 0.4574                               \\ \hline
\multicolumn{1}{c}{}        & \multicolumn{1}{c}{\textbf{\textit{MIXAD}}} & \multicolumn{1}{c}{0.9703}    & \multicolumn{1}{c}{0.9884} & \multicolumn{1}{c|}{\underline{0.9792}} & \multicolumn{1}{c}{\textbf{0.5137}} & \multicolumn{1}{c}{\textbf{0.6672}} \\ \hline
                              & MTAD-GAT                          & 0.9919                         & 0.7964                      & 0.8835                            & \underline{0.5812}                         & 0.5885                               \\
                              & GDN                               & 0.9989                        & 0.8026                      & 0.8900                            & 0.2276                               & 0.3382                               \\
MSDS                          & TranAD                            & 0.9859                         & 0.9749                      & \textbf{0.9804}                   & 0.4583                               & 0.6253                               \\
                              & DuoGAT                            & 0.9634                         & 0.9576                      & 0.9605                            & 0.4435                               & \underline{0.6614}                         \\
                              & DAEMON                            & 0.9711                         & 0.9450                      & 0.9578                            & 0.3358                               & 0.5115                               \\ \hline
\multicolumn{1}{c}{}        & \multicolumn{1}{c}{\textbf{\textit{MIXAD}}} & \multicolumn{1}{c}{0.9716}    & \multicolumn{1}{c}{0.9540} & \multicolumn{1}{c|}{\underline{0.9627}}       & \multicolumn{1}{c}{\textbf{0.7818}} & \multicolumn{1}{c}{\textbf{0.8136}} \\ \hline
\end{tabular}
}
\label{Tab:table2}
\end{table*}
\end{center}

\subsection{Performance Comparisons}
We thoroughly evaluated \textit{MIXAD}'s performance through comprehensive experiments, with setup details provided in the supplementary materials. Utilizing a grid search method, we determined the optimal anomaly thresholds for each experiment based on the highest F1-scores~\cite{OmniAnomaly}. As shown in Table~\ref{Tab:table2}, \textit{MIXAD} achieved competitive detection accuracy for both the SMD and MSDS datasets, closely matching SOTA models. Notably, while its detection accuracy in the SMD dataset ranked second, slightly below the top SOTA model by 1.73\% and above the second-best by 2.05\%, \textit{MIXAD} significantly outperformed all models in interpretability. It improved interpretation scores by \textbf{34.30\%} and \textbf{29.43\%} in HitRate@100\% and HitRate@150\%, respectively, surpassing the previously highest-ranking DuoGAT algorithm. In the MSDS dataset, \textit{MIXAD} again showed detection performance slightly below the best by 1.81\%, yet it improved upon the second highest by 0.23\%, placing it second overall. More impressively, it raised the bar for interpretability, setting new records with increases of \textbf{34.51\%} and \textbf{23.01\%} in HitRate@100\% and HitRate@150\%, respectively. These highlight \textit{MIXAD}'s strong detection abilities and its superior interpretative performance, demonstrating its potential for real-world applications.

\begin{center}
    \begin{table*}[!t]
\centering
\caption{Ablation study on the SMD dataset. The highest performance according to the F1-score and HitRate@P\% is highlighted in bold.}
\scalebox{0.9}{
\begin{tabular}{cccccc}
\hline
\multicolumn{1}{c}{\textbf{Methods}}      & \multicolumn{1}{c}{\textbf{Precision}} & \multicolumn{1}{c}{\textbf{Recall}} & \multicolumn{1}{c}{\textbf{F1}}     & \multicolumn{1}{c}{\textbf{HitRate@100\%}} & \multicolumn{1}{c}{\textbf{HitRate@150\%}} \\ \hline
- Reconstruction                            & 0.9410                                  & 0.9864                               & 0.9629                               & 0.4130                                      & 0.5562                                      \\
- $L_3$                                      & 0.9617                                  & 0.9719                               & 0.9667                               & 0.3537                                      & 0.4868                                      \\
- Memory module                             & 0.9793                                  & 0.9493                               & 0.9626                               & 0.4317                                      & 0.5695                                      \\
- New anomaly score                         & 0.9767                                  & 0.9506                               & 0.9624                               & 0.4278                                      & 0.5658                                      \\
- New interpretation                        & 0.9703                                  & 0.9884                               & \textbf{0.9792}                      & 0.4829                                      & 0.5970                                      \\ \hline
\multicolumn{1}{c}{\textbf{\textit{MIXAD}}} & \multicolumn{1}{c}{0.9703}             & \multicolumn{1}{c}{0.9884}          & \multicolumn{1}{c}{\textbf{0.9792}} & \multicolumn{1}{c}{\textbf{0.5137}}        & \multicolumn{1}{c}{\textbf{0.6672}}        \\ \hline
\end{tabular}
}
\label{Tab:table3}
\end{table*}
\end{center}

\subsection{Ablation Study}
Our ablation study aimed to evaluate the individual contributions of various components within \textit{MIXAD} towards its detection and interpretation capabilities. The study involved several modifications: (1) replacing the reconstruction-focused decoder with a forecasting one, (2) removing the KL divergence loss ($L_3$) which promotes uniform memory activation, (3) substituting the memory module ($M$) with a learnable node embedding similar to the GDN baseline~\cite{GDN}, (4) discarding the newly proposed anomaly scoring method and instead relying solely on reconstruction error for scoring, and (5) excluding our novel interpretation method based on Pearson correlation among anomaly scores. The results are detailed in Table~\ref{Tab:table3}. The findings highlight the integral role each component plays in \textit{MIXAD}'s performance. Notably, the exclusion of any key component diminished both detection and interpretability, illustrating their collective importance. However, the omission of the novel interpretation method did not affect detection accuracy, as it employed the same enhanced anomaly scoring mechanism $s'$. Our analysis indicates that while all components significantly enhance interpretability, the KL divergence loss ($L_3$) is particularly effective. It prevents biased learning that could lead to the underutilization of some memory slots, thereby markedly improving model explainability. Additionally, the proposed post hoc interpretation technique plays a complementary role, further refining the final interpretability of the detection output.

\begin{figure*}[!t]
    \centering
    \begin{minipage}{0.58\textwidth}
        \centering
        \includegraphics[width=\linewidth]{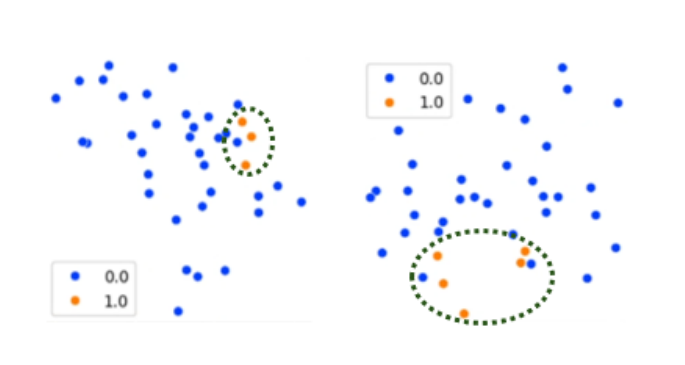}
        \caption{T-SNE visualization of node embeddings from two anomaly segments of the SMD dataset.}
        \label{fig:TSNE}
    \end{minipage}\hfill
    \begin{minipage}{0.38\textwidth}
        \centering
        \includegraphics[width=\linewidth]{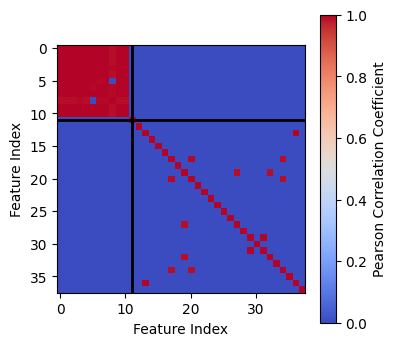}
        \caption{Heatmap visualization of Pearson correlation coefficients for anomaly scores.}
        \label{fig:Correlation}
    \end{minipage}
\end{figure*}

\subsection{Visualization of Node Embeddings}
We conducted a qualitative assessment of the node embeddings' quality by employing t-SNE to visualize them in a low-dimensional space. As depicted in Figure~\ref{fig:TSNE}, the orange and blue points represent the root cause and non-causal features of two anomalous segments within the SMD dataset, respectively. Consistent with our expectations, the visualization reveals that the memory-based node embeddings form distinct clusters based on feature relationships and their contributions to anomalies. This clustering demonstrates the memory module's ability to capture and retain each feature's unique spatiotemporal attributes, allowing for the creation of analogous embeddings for similar nodes. Such capability not only aids in accurate time series reconstruction but also significantly boosts the model's effectiveness in anomaly detection and interpretation, showcasing the memory's integral role in improving \textit{MIXAD}'s functionality.

\begin{figure*}[!t]
 \centering
 \includegraphics[width = 0.793\textwidth]{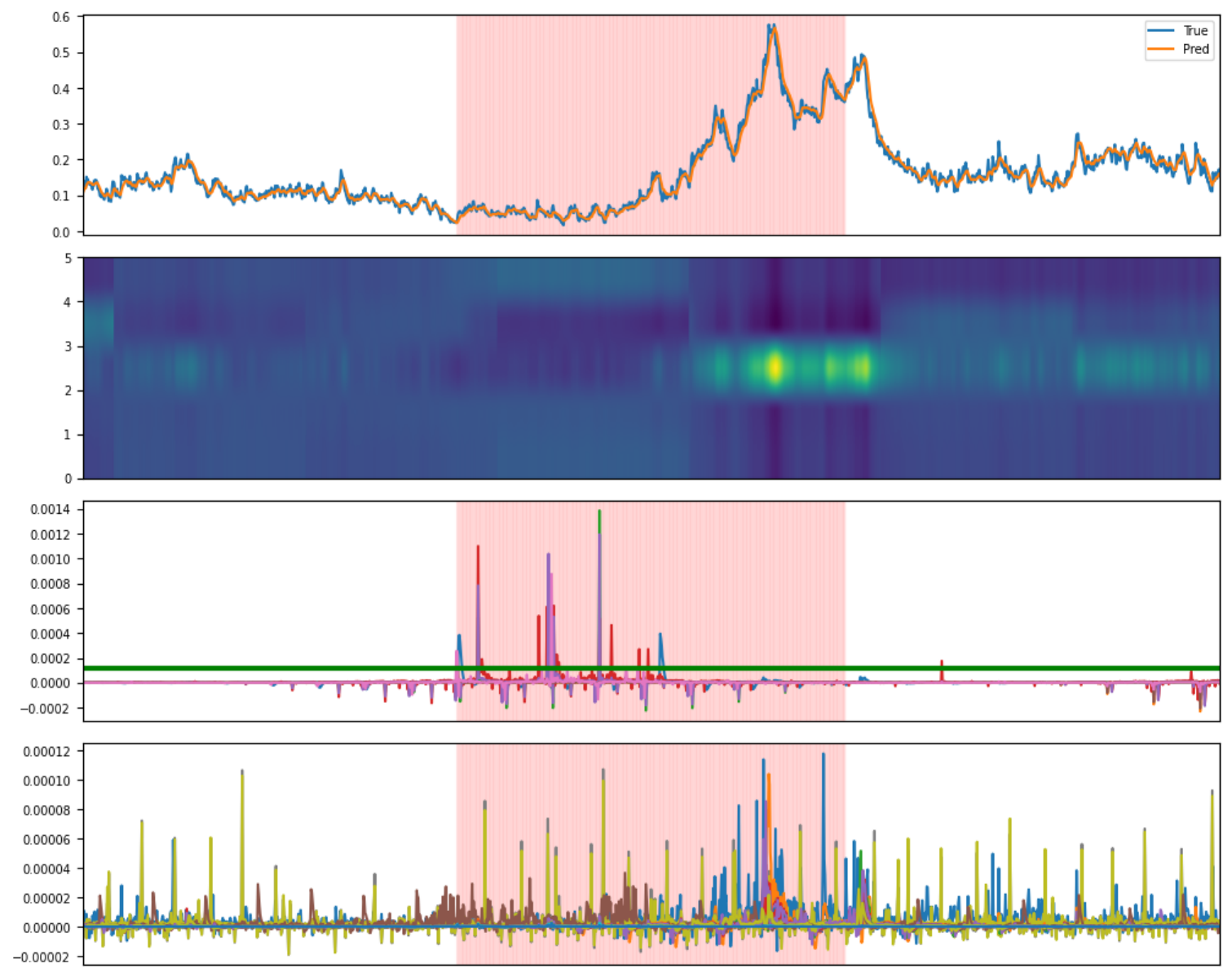}
 \caption{Visualization of memory activation and anomaly scores for an anomaly segment in the SMD dataset.}
 \label{fig:Scores}
\end{figure*}

\subsection{Visualization of Anomaly Scores}
To demonstrate the efficacy of \textit{MIXAD}'s innovative anomaly scoring method, we visualized an anomalous segment from the SMD dataset, depicted in Figure~\ref{fig:Scores}. This figure consists of four sequential graphs. The top graph illustrates the original time series in blue alongside its reconstruction in orange. The second graph shows the memory activation maps (attention scores) across timestamps, followed by two graphs that display anomaly scores for causal and non-causal features, respectively. A green horizontal line in the third graph highlights the maximum anomaly score among the non-causal features, and a red shaded area across all graphs marks the anomaly segment's duration. This visualization reveals a discernible shift in memory attention within the anomalous period despite the close resemblance between the actual and reconstructed time series. Importantly, anomaly scores based on memory attention are significantly elevated only for the causal features within this segment. This clear differentiation supports the effectiveness of our anomaly scoring approach in accurately identifying and interpreting anomalies, emphasizing its potential utility.

Furthermore, we analyze anomaly score correlations within a specific anomalous segment in the SMD dataset, as shown in Figure~\ref{fig:Correlation}. We calculate the Pearson correlation coefficient for anomaly scores ${s'}_{seg}$ across nodes over the duration of an anomaly segment $seg$, generating an $N \times N$ correlation matrix. This matrix is visually depicted as a heatmap, where the top-left box, divided by horizontal and vertical black lines, illustrates the correlation among the root cause features. Notably, features responsible for the anomaly demonstrate a high correlation in their scores, underlining the similar memory activation shift each of these causal features exhibits. Thus, this correlation coefficient serves as a basis for facilitating more accurate interpretations.

\begin{table*}[!t]
\centering
\caption{Anomaly detection performance evaluation on the Exathlon dataset.}
\scalebox{0.9}{
\begin{tabular}{c|r|r}
\hline
\textbf{Model}        & \multicolumn{1}{c|}{\textbf{AD1(F1-score)}} & \multicolumn{1}{c}{\textbf{AD2(F1-score)}} \\ \hline
\textbf{TranAD}       & 0.2166                              & 0.2166                             \\
\textbf{DuoGAT}       & \textbf{0.9900}                              & \underline{0.1296}                             \\ \hline
\textbf{\textit{MIXAD}} & \underline{0.9665}                              & \textbf{0.1526}                             \\ \hline
\end{tabular}
}
\label{Tab:table3}
\end{table*}

\section{Case Study: Exathlon Dataset \& Testbed}
\begin{figure*}[!t]
 \centering
 \includegraphics[width = 0.85\textwidth]{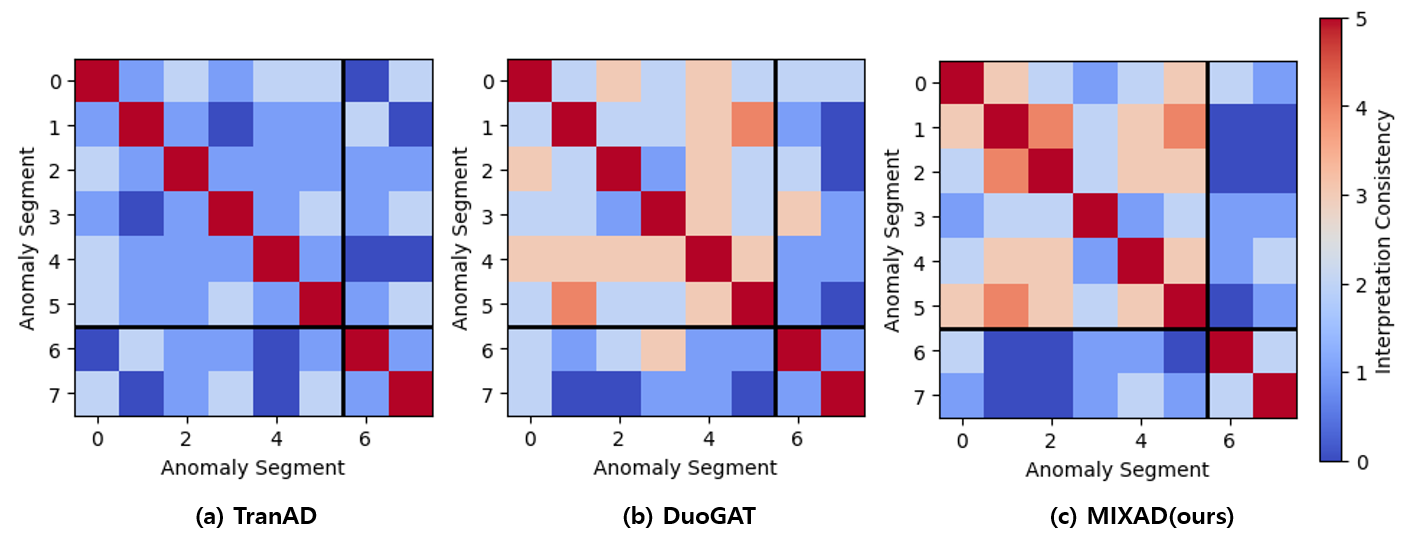}
 \caption{Anomaly interpretation performance evaluation on the Exathlon dataset.}
 \label{fig:exathlon}
\end{figure*}

To evaluate our model's efficacy on a real-world benchmark, we used the Exathlon dataset and testbed from~\cite{exathlon}, comparing MIXAD with the top baselines DuoGAT and TranAD. Exathlon, unlike the small benchmarks used in Section~\ref{section:experiment}, has 2,283 dimensions and includes noisy training data for a realistic scenario. We experimented with data from Spark streaming application 1, containing two types of anomalies, each with 6 and 2 segments, respectively. The testbed in~\cite{exathlon} supports range-based evaluation, which is not directly applicable to point-based evaluation. Thus, we used a point-based version of AD1 and AD2 metrics from~\cite{exathlon}. AD1 is equivalent to our point-adjusted evaluation, while AD2 is non-point-adjusted. Table~\ref{Tab:table3} shows that TranAD fails completely, flagging all timestamps as anomalous. DuoGAT and MIXAD detect anomalies with high f1-scores over 0.95 in AD1, but their performance drops in AD2, indicating reliance on point-adjustment. MIXAD, designed to detect anomaly segments using memory activation shifts, still achieves a higher f1-score in AD2, proving its superiority. Furthermore, MIXAD's explanations include which features caused anomalies, while Exathlon lacks ground truth feature labels. Therefore, we modified the consistency metric from~\cite{exathlon}. Explanations for the same anomaly type should be similar, while those for different types should differ. We extracted the top-5 causal features for each detected anomaly and quantified consistency by counting intersections between segment pairs. In Figure~\ref{fig:exathlon}, the six anomaly segments in the upper left are type 1, and the two in the lower right are type 2. Figure~\ref{fig:exathlon} shows that MIXAD provides the most consistent explanations for the same type, proving its superiority in interpreting anomalies in the Exathlon dataset.

\section{Conclusion}
In this paper, we introduce \textit{MIXAD}, an interpretable MTS anomaly detection model designed to effectively capture and store node-level prototypes of fine-grained spatiotemporal patterns. Leveraging the STRGC framework and a memory augmentation method, \textit{MIXAD} offers a more accountable approach to understanding complex data relationships. Our novel anomaly scoring technique, which utilizes memory activation pattern analysis, significantly improves interpretability in MTS anomaly detection. While \textit{MIXAD} may not achieve SOTA results in anomaly detection, its interpretability and the empirical insights it provides mark important advancements for future research in the field.

\subsubsection{Acknowledgements.} 
This work was partly supported by Institute for Information \& communication Technology Planning \& evaluation (IITP) grants funded by the Korean government MSIT:
(RS-2022-II221199, RS-2024-00337703, RS-2022-II220688, RS-2019-II190421, RS-2023-00230337, RS-2024-00356293, RS-2022-II221045, and RS-2021-II212068).


\bibliographystyle{abbrv}
\bibliography{mybibfile}
\end{document}


\title{Supplementary Materials}

\author{}
\authorrunning{}
\institute{}
\maketitle

\section{Recent Advances in Multivariate Anomaly Detection}
Multivariate anomaly detection methods in time series analysis are broadly classified into forecasting-based and reconstruction-based approaches. Forecasting-based methods use prediction errors for anomaly detection. Notably, Hundman et al.\cite{hundman2018detecting} introduce an unsupervised, non-parametric thresholding method to interpret predictions from Long Short-Term Memory (LSTM) networks for spacecraft telemetry monitoring. Similarly, Ding et al.\cite{ding2018multivariate} propose a real-time detection algorithm leveraging Hierarchical Temporal Memory and Bayesian Networks, while Gugulothu et al.\cite{gugulothu2018sparse} combine dimension reduction techniques with recurrent autoencoders in an end-to-end framework for time series modeling. Zong et al.\cite{zong2018deep} address the detection of anomalies in multivariate data without temporal dependencies by analyzing observations across multiple dimensions. Recent methods also utilize advanced architectures like graph neural networks to consider inter-metric relations. Deng et al.\cite{GDN} employ graph neural networks and sensor embeddings to analyze complex sensor relationships in multivariate time series, aiming to improve detection accuracy and interpretability. Lee et al.\cite{DuoGAT} expand this approach by integrating dual graph attention networks, further enhancing the model's ability to capture temporal nuances and significantly boosting detection efficiency.

Reconstruction-based anomaly detection models learn representations of entire time series by reconstructing the original input from latent variables. Malhotra et al.\cite{EncDec-AD} introduce an LSTM-based Encoder-Decoder framework for learning from normal time series for anomaly detection. Mirsky et al.\cite{mirsky2018kitsune}, employing an unsupervised approach, use autoencoders to reconstruct instance features from integrated visible neurons. Park et al.\cite{park2018multimodal} combine LSTM with variational autoencoders to fuse signals and reconstruct the expected distribution of multivariate observations, addressing temporal dependencies through both encoding and decoding processes. Generative Adversarial Networks (GANs) are also widely used in multivariate time series anomaly detection. Li et al.\cite{li2019mad} explore latent interactions among variables by considering the entire variable set, while Li et al.\cite{li2018anomaly} introduce a novel GAN-based method that employs the discriminator and residuals between the generator-reconstructed data and actual samples for anomaly detection. Recent advancements include the adoption of sophisticated architectures to enhance detection outcomes. Zhao et al.\cite{MTAD_GAT} employ parallel graph attention layers to refine anomaly detection by optimizing both forecasting and reconstruction efforts, thus improving the model's representation and anomaly identification capabilities. Tuli et al.\cite{TranAD} utilize transformer networks with attention-based encoders and adversarial training to enhance detection speed and accuracy. Chen et al.\cite{DAEMON}, leveraging adversarial autoencoders in an unsupervised context, use dual discriminators and reconstruction errors for anomaly detection, offering robustness and improved interpretability through gradient analysis.

\begin{center}
    \begin{table*}[!t]
\centering
\caption{Dataset Statistics}
\resizebox{ 1 \textwidth }{!}{%
\begin{tabular}{c|r|r|r|r|r}
\hline
\textbf{Dataset} & \multicolumn{1}{c|}{\textbf{Subsets}} & \multicolumn{1}{c|}{\textbf{Dimensions}} & \multicolumn{1}{c|}{\textbf{Training set size}} & \multicolumn{1}{c|}{\textbf{Testing set size}} & \multicolumn{1}{c}{\textbf{Anomaly ratio (\%)}} \\ \hline
SMD     & 4                            & 38                              & 708405                                 & 708420                                & 4.16                                    \\ \hline
MSDS    & 1                            & 10                              & 146430                                 & 146430                                & 5.37                                    \\ \hline
\end{tabular}}
\label{Tab:table1}
\end{table*}
\end{center}

\section{Dataset Details}
\begin{enumerate}
    \item \textbf{Server Machine Dataset (SMD)~\cite{OmniAnomaly}:} This dataset comprises five weeks of detailed resource utilization traces from 28 machines within a computing cluster. Notably, the dataset contains sequences that might be considered trivial for anomaly detection purposes~\cite{Trivial}. Therefore, we specifically focus on the non-trivial sequences within this dataset, including the traces from machines 1-1, 2-1, 3-2, and 3-7. These selected traces provide a richer source of data for testing the effectiveness of our anomaly detection approach.
    \item \textbf{Multi-Source Distributed System (MSDS) Dataset~\cite{MSDS}:} Representing a more recent addition to available resources, the MSDS dataset encompasses a diverse array of data types, including distributed traces, application logs, and system metrics, from a complex distributed system. Designed with AI operations in mind, this dataset serves as an ideal testing ground for automated anomaly detection, root cause analysis, and remediation tasks. Its multi-source nature and focus on high-quality, actionable data make it particularly well-suited for evaluating the performance of advanced anomaly detection and interpretation methodologies.
\end{enumerate}
These datasets play a crucial role in our experiments, enabling us to rigorously evaluate and showcase the improved interpretability of anomaly detection achieved through our proposed method. We summarize the dataset statistics in Table~\ref{Tab:table1}.

\section{Baselines}
\begin{enumerate}
    \item \textbf{MTAD-GAT~\cite{MTAD_GAT}:} This model integrates parallel graph attention layers to analyze temporal dynamics and inter-feature connections. It aims to refine anomaly detection by jointly optimizing both forecasting and reconstruction efforts, thereby enhancing the model's capability to represent and accurately identify anomalies.
    \item \textbf{GDN~\cite{GDN}:} Utilizing graph neural networks combined with sensor embeddings, GDN is designed to uncover intricate relationships among sensors within multivariate time series. This approach aims to boost both the accuracy of anomaly detection and the clarity with which these detections can be interpreted.
    \item \textbf{TranAD~\cite{TranAD}:} Leveraging the capabilities of transformer networks, TranAD advances anomaly detection in multivariate time series data. It employs attention-based encoders, adversarial training, and self-conditioning techniques to not only expedite detection but also to increase the accuracy of its predictions, ensuring effective and efficient training processes.
    \item \textbf{DuoGAT~\cite{DuoGAT}:} By integrating dual graph attention networks within a unique graph structure, DuoGAT excels in capturing the temporal nuances essential for anomaly detection in multivariate time series. This method is notable for its ability to significantly elevate both the performance and efficiency of detection tasks.
    \item \textbf{DAEMON~\cite{DAEMON}:} DAEMON makes use of adversarial autoencoders in an unsupervised learning context to detect anomalies in multivariate time series. It distinguishes itself by employing dual discriminators to refine the identification of normal patterns and utilizing reconstruction errors as a basis for anomaly detection. This model is recognized for its robustness and offers enhanced interpretability, particularly through the application of gradient analysis in anomaly diagnosis.
\end{enumerate}

\section{Experimental Setup}
Our proposed \textit{MIXAD} model was developed using PyTorch version 1.12.1 and CUDA version 11.2, and our experiments were conducted on a server equipped with an Intel(R) Xeon(R) Silver 4114 CPU @ 2.20GHz and four TITAN RTX GPUs. In alignment with the recommendations from the original studies, we utilized a window length $w$ of 30 for both the SMD and the MSDS. The datasets collected under normal conditions were divided into training and validation sets in an 8:2 ratio. For the construction of the memory module $M$, we designated $m=5$ for SMD and $m=2$ for MSDS, with both configurations sharing a uniform dimensionality $d$ of 64. Both the encoder and decoder were configured to employ a single layer of STRGC, utilizing Chebyshev polynomials of degrees $K=3$ for SMD and $K=2$ for MSDS, with a consistent hidden representation size $h$ of 64. The loss component coefficients were set to $\lambda_1 = 0.01,0.001$, $\lambda_2 = 0.1,0.001$, $\lambda_3 = 0.0001,0.0001$ for SMD and MSDS, respectively. Training was performed with a batch size of 256 using the Adam optimizer at a learning rate of 0.001 across 30 epochs, incorporating an early stopping mechanism triggered after 10 epochs without improvement.

\bibliographystyle{abbrv}
\bibliography{mybibfile}